\documentclass{article}

\usepackage{microtype}
\usepackage{graphicx}
\usepackage{subcaption}
\usepackage{booktabs} 

\usepackage{hyperref}

\usepackage[preprint]{icml2026}

\usepackage{amsmath}
\usepackage{amssymb}
\usepackage{mathtools}
\usepackage{amsthm}

\usepackage[capitalize,noabbrev]{cleveref}

\theoremstyle{plain}

\theoremstyle{definition}

\theoremstyle{remark}

\usepackage[disable,textsize=tiny]{todonotes}

\icmltitlerunning{Phase-Aware Wavelet-Based-Scattering Encoder-Decoder}

\begin{document}

\twocolumn[
  \icmltitle{Phase-Aware Wavelet-Based-Scattering Encoder-Decoder for Dense Predictions}




  \begin{icmlauthorlist}
    \icmlauthor{Ghassen MARRAKCHI}{uspn}
    \icmlauthor{Basarab MATEI}{uspn}
  \end{icmlauthorlist}

  \icmlaffiliation{uspn}{Northern Paris Computer Science Lab, Sorbonne Paris Nord University, Villetaneuse, France}

  \icmlcorrespondingauthor{Ghassen MARRAKCHI}{ghassen.marrakchi@lipn.univ-paris13.fr}
  \icmlcorrespondingauthor{Basarab MATEI}{matei@lipn.univ-paris13.fr}

  \icmlkeywords{Deep Learning, Scattering, Dense, Phase, Computer Vision}

  \vskip 0.3in
]



\printAffiliationsAndNotice{}  

\begin{abstract}
Scattering transforms achieve Lipschitz stability and translation invariance, but dense prediction tasks require preserving spatial structure lost in global averaging. We propose Phase-Aware Scattering Encoder-Decoder, which restores this information by explicitly preserving phase in skip connections. On image denoising (BSD68), breaking translation invariance improves PSNR by +2.17~dB; phase preservation adds +1.03~dB. A novel spatial shuffling ablation ($-1.26$~dB penalty) demonstrates phase encodes location-dependent structure. We conduct a preliminary extensibility study on a second dense prediction task (ISIC skin lesion segmentation), with full cross-validation as ongoing work. This work advances principled wavelet-deep learning integration, showing how phase information complements scattering's stability-expressiveness trade-off in pixel-level prediction.
\end{abstract}

\section{Introduction}
\label{sec:intro}
Scattering transforms \cite{mallat2012group, bruna2013invariant} provide mathematically-grounded signal representations through cascaded wavelet-modulus operations, achieving translation invariance and Lipschitz stability---valuable properties for classification. However, scattering's core mechanism---global averaging and downsampling by $2^J$---fundamentally conflicts with dense prediction tasks (denoising, segmentation) where pixel-level spatial correspondence is essential. Standard scattering compresses $H \times W$ images to $H/2^J \times W/2^J$ representations, discarding the fine spatial information required for pixel-accurate reconstruction.

Scattering achieves two distinct properties via different mechanisms. Lipschitz stability derives from cascaded modulus operations (non-expansive nonlinearities), which remain valid regardless of spatial resolution. Translation invariance derives from downsampling and global averaging, which are not essential to stability---only to invariance.

While global translation invariance from subsampled averaging is theoretically elegant and valuable for classification, it conflicts with dense prediction tasks, where pixel-level spatial correspondence is essential. The observed decoupling suggests a possibility: can we preserve stability while recovering spatial equivariance for dense prediction?

We propose a phase-aware scattering encoder-decoder architecture that breaks global invariance (via stride-1 processing and per-scale smoothing) while preserving Lipschitz stability and adding explicit phase preservation. This design enables spatial structure recovery while maintaining mathematical interpretability and robustness guarantees---providing an interpretable alternative to black-box learned methods for safety-critical applications (medical imaging, denoising).

\section{Related Work}
\label{sec:related}
Encoder-decoder architectures have become the standard paradigm for dense prediction tasks. U-Net \cite{ronneberger2015} established skip connections for preserving spatial information during upsampling, becoming the standard paradigm for segmentation and denoising. Recent extensions employ vision transformers (\textit{TransUNet} \cite{CHEN2024103280}, \textit{Swin-UNet} \cite{cao2022swin}) replacing convolutional encoders with hierarchical attention, achieving state-of-the-art results. In denoising specifically, specialized architectures like \textit{DnCNN} \cite{7839189} and \textit{SwinIR} \cite{9607618} have demonstrated superior performance through task-adapted designs. Our approach replaces learned convolutional encoders with mathematically-grounded scattering wavelets while augmenting skip connections to preserve phase information---combining multi-scale context with interpretability.

Other work integrates wavelets into deep learning: \textit{MWCNN} \cite{liu2018multi} decomposes images into wavelet levels for independent processing; \textit{SDWNet} \cite{zou2021sdwnet} preserves full spatial resolution via dilated convolutions. Beyond magnitude-only approaches, theory \cite{mallat2020phase} and empirics \cite{Cole2021-oz,DRAMSCH2021104643} demonstrate that magnitude information alone cannot capture phase full structural content---edges, singularities---essential for reconstruction, and complex-valued networks outperform real-valued. Our phase-aware scattering encoder-decoder validates this insight: complex-valued skip connections enable spatial structure recovery while maintaining scattering's Lipschitz stability guarantees.

Recent work has begun addressing phase preservation within scattering itself. Wang et al. (\citeyear{wangcomplex}) propose H-CReLU, a learnable activation functions that preserve phase within scattering nonlinearities. Our approach is complementary: rather than modifying scattering's computation, we preserve amplitude and phase information through carefully designed encoder-decoder skip connections that explicitly route complex-valued scattering coefficients to reconstruction layers.

We address gaps across four dimensions: (1) extend scattering from classification to dense prediction via stride-1 equivariance, (2) integrate phase preservation into mathematically-grounded scattering architecture, (3) combine cascaded nonlinear wavelets with phase-aware skip connections, (4) provide interpretable alternative to black-box architecture search and transformer opacity.

\section{Background}
\label{sec:back}

The scattering transform \cite{mallat2012group, bruna2013invariant} constructs a stable, informative signal representation by cascading wavelet convolutions with modulus nonlinearities. It proceeds through three stages: (1) wavelet convolution producing complex coefficients $W[\lambda]x = x \star \psi_\lambda$, (2) modulus nonlinearity yielding real coefficients $U[\lambda]x = |W[\lambda]x|$, and (3) averaging and downsampling producing final coefficients $S_J[p]x = (U[p]x \star \phi_J)(2^J u)$.

This design yields three fundamental properties:

\paragraph{Translation Invariance:}
Global averaging and downsampling by $2^J$ ensure $S_J[T_\tau x] = S_J[x]$ for translations $|\tau| < 2^J$. Critical insight: subsampling is the mechanism achieving invariance; averaging alone does not suffice.

\paragraph{Lipschitz Stability:}
Cascaded modulus operations are non-expansive (Lipschitz constant 1), so $|Sx - Sy| \leq |x - y|$. This property linearizes small diffeomorphisms and provides robustness guarantees.

\paragraph{Energy Decay:}
Propagator energy decreases exponentially with path length, permitting truncation at low orders (typically $m=2$) while retaining most signal energy. This ensures minimal information loss-- better expected information reconstruction.

These properties matter for dense prediction: Invariance is prohibitive (discards spatial information), but stability and decay are valuable. Our modifications (stride-1 \& phase preservation) preserve stability and decay while breaking invariance to recover spatial structure (Section~\ref{sec:method}).

\section{Methodology}
\label{sec:method}

\begin{figure*}[t]
  \vskip 0.2in
  \begin{center}
    \centerline{\includegraphics[width=\textwidth]{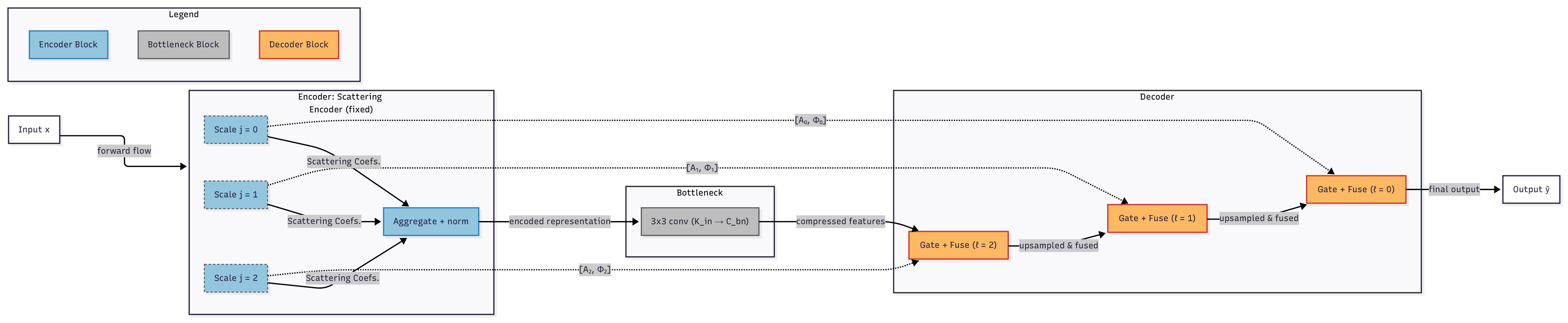}}
    \caption{
      \textbf{Phase Aware Wavelet-Based Scattering Encoder-Decoder:} input $\rightarrow$ scattering encoder (3 scales, 8 orientations, stride-1) $\rightarrow$ aggregation and normalization $\rightarrow$ bottleneck (single 3×3 conv) $\rightarrow$ 3 decoder levels with phase-aware gating and fusion $\rightarrow$ output, with skip connection paths for ($A_j$, $\Phi_j$) explicitly labeled
    }
    \label{fig:full architecture-fig}
  \end{center}
\end{figure*}

Our phase-aware scattering-based U-Net-like (Encoder, Bottleneck, Decoder) architecture (depicted in figure~\ref{fig:full architecture-fig}) combines two key architectural innovations to enable dense prediction while preserving mathematical interpretability.

\paragraph{Stride-1 Equivariance (Removing Subsampling):} 
We remove global averaging at scale $2^J$ used in Standard scattering, and maintain full spatial resolution throughout the scattering hierarchy. Each scale receives localized smoothing windows $\phi_j$ proportional to $2^j$ (per-scale). This preserves the critical equivariance property:
\begin{equation}
\begin{split}
\Phi(L_\tau x) = L_\tau \Phi(x)\\
\end{split}
\label{eq:method-eq-1}
\end{equation}
where $L_\tau$ is translation by $\tau$ pixels. Stride-1 processing ensures the bottleneck receives fully resolved spatial structure, enabling pixel-accurate reconstruction. The trade-off is explicit: we sacrifice global translation invariance, but Lipschitz stability (derived from cascaded modulus operations) is preserved.

\paragraph{Phase Preservation (Complex-Valued Skip Connections):}
We explicitly preserve, originally discarded, complex-valued wavelet coefficients and extract both magnitude and phase:
\begin{align}
    A[p]x(u) &= |W[\lambda]x(u)|, \label{eq:method-magnitude} \\
    \Phi[p]x(u) &= \arg(W[\lambda]x(u)). \label{eq:method-phase}
\end{align}

Phase information encodes location-dependent structural detail: edges, singularities, and boundary localization that magnitude alone cannot represent. Theoretically, phase harmonics characterize coherent structures via phase correlations \cite{mallat2020phase}. Empirically, we show phase is $5.14\times$ more important than amplitude for reconstruction (Section~\ref{sec:exp}).

Rather than keeping Cartesian form (real, imaginary), skip connections are made in polar form $(A_j, \Phi_j)$. This decouples phase and magnitude gradients during backpropagation, isolating phase as an independent learned signal. Phase carries $\sim$90\% of structural content in natural images \cite{nie2021neural}; polar form prioritizes phase-driven geometric details without interference from magnitude gradients.

\subsection{Encoder: Scattering Transform with Stride-1 and Phase Preservation}
\label{subsec:method-encoder}
The encoder (depicted in figure~\ref{fig:encoder-fig}) is a modified scattering transform with fixed wavelet filters. It retains mathematical stability guarantees while replacing standard averaging with scale-dependent smoothing to preserve full spatial dimensions. 
For an input image $x \in \mathbb{R}^{C \times H \times W}$, the encoder computes:

\paragraph{Multi-scale Decomposition:}
Cascaded wavelet decomposition with $J$ scales and $L$ orientations. At each scale $j$ and orientation $\theta$, we compute:
\begin{align}
    W_j^\theta x &= x \ast \psi_j^\theta \label{eq:method-encoder-eq-1}
\end{align}
where $\ast$ denotes convolution and $\psi_j^\theta$ are directional Gabor-like wavelets. All coefficients are computed at full spatial resolution (stride 1, no subsampling).

We employ complex Morlet wavelets from the Kymatio library \footnote{Under the MIT License (© 2022 Kymatio Team).} \cite{andreux2022kymatio} as the directional filter bank $\psi^\theta_j$ — a deliberate design choice, as complex-valued wavelets produce the imaginary component from which phase is extracted via Eq. \ref{eq:method-encoder-eq-4}. This makes complex wavelet selection a formal architectural prerequisite of the phase preservation mechanism.

This is performed in a two-order fashion, with a global low-pass reference resolution, whose coefficients are defined as follows:
\begin{align}
    \text{(Reference-order)} \quad S_0 &= x \ast \phi_J, \label{eq:method-encoder-s0} \\
    \text{(First-order)} \quad S_1^{(j)} &= |W_j^\theta x| \ast \phi_j, \label{eq:method-encoder-s1} \\
    \text{(Second-order)} \quad S_2^{(j)} &= \big||W_j^\theta x| \ast \psi_j^\theta\big| \ast \phi_j, \label{eq:method-encoder-s2}
\end{align}
where $S_1$ is computed across all orientations and $S_2$ accounts for sibling paths.

\paragraph{Aggregation:}
All coefficients (low-pass, first-order, second-order) are collected into a unified tensor, organized by scale $j$ (considered by the decoder, in the U-Net architecture terminology, as level $\ell$). Orientations are flattened into the channel dimension, creating a descriptor of multi-scale structure. Total channels: 
\begin{align}
    K_{\text{in}} &= C_{\text{in}} \left(1 + J \cdot L + \sum_{j=0}^{J-1} K_{\text{siblings}}(j) \cdot L^2\right) \label{eq:method-encoder-eq-2}
\end{align}

Aggregated features are then normalized by a factor ensuring total energy matches the low-pass component (Frobenius norm scaling).

\paragraph{Complex-Valued Phase Extraction:}
Before final modulus operations, pre-modulus complex coefficients are stored separately and converted to polar form (magnitude + phase):
\begin{align}
    A_j^\theta &= \sqrt{(\text{Re}(W_j^\theta))^2 + (\text{Im}(W_j^\theta))^2 + \epsilon}, \label{eq:method-encoder-eq-3}\\
    \Phi_j^\ell &= \text{atan2}(\text{Im}(W_j^\theta), \text{Re}(W_j^\theta)), \label{eq:method-encoder-eq-4}
\end{align}
where $\epsilon = 10^{-6}$ provides numerical stability, and $\text{atan2} \in [-\pi, \pi]$.

We get, at each scale $j$: $\mathcal{E}_j = (A_j, \Phi_j)$

\paragraph{Output:} (i) The aggregated scattering coefficients (normalized, $K_{\text{in}}$ channels), and (ii) the phase information from all scales (stored separately for decoder skip connections)

\subsection{Bottleneck: Learned Dimensionality Reduction Bridge}
\label{subsec:method-bottleneck}

The bottleneck forms a principled bridge between the fixed scattering encoder and learned decoder. Having the aggregated scattering tensor with $K_{\text{in}}$ channels (can exceed 1000+ for $J=3$), it reduces computational complexity, by applying a bottleneck transformation $\Psi_{\text{bn}}$ consisting of a single $3\times3$ convolution with stride $1$:
\begin{equation}
    \mathbf{Z} = \Psi_{\text{bn}}(\mathbf{F}), \label{eq:bottleneck}
\end{equation}
where $\mathbf{F}$ is the input feature volume. This layer maps the signal from $K_{\text{in}}$ to $C_{\text{bn}}$ channels (e.g., $C_{\text{bn}}=64$ for BSD68).

The bottleneck goal is to compress high-dimensional scattering representations into a practical latent code. It learns task-relevant nonlinear combinations without excessive learnable complexity, and maintains full spatial resolution: $Z \in \mathbb{R}^{(B \times C_{\text{bn}} \times H \times W)}$ preserves encoder resolution and ensures spatial equivariance for pixel-level reconstruction.

\paragraph{Output:} Compact latent representation with preserved spatial structure, fed to decoder stages. Profiling on $128 \times 128$ inputs shows M6 (Ours) uses $2\times$ fewer parameters and FLOPs than DnCNN, and $172\times$ fewer parameters than DSWN, with zero learnable parameters in the scattering encoder.

\subsection{Decoder: Phase-Aware U-Net-like with Learned Gating and Fusion}
\label{subsec:method-decoder}

The decoder progressively upsamples bottleneck features while fusing multi-scale phase-guided skip connections. It is a $J$ upsampling levels (one per scattering scale), iterating from coarse to fine. Each level processes three information sources:

\paragraph{Skip Connections (Phase-Guided):} At each decoder level $\ell$, the corresponding scattering scale contributes pre-modulus complex coefficients in polar form $(A_\ell, \Phi_\ell)$, with the shape: $(B, C_{\text{in}}\cdot L, H_\ell, W_\ell)$ for each of magnitude and phase.

\paragraph{Gating Network (Phase-Informed Attention):} Building on recent phase-aware gating approaches \cite{Chen2025, GUO2025103201}, we use a phase gating mechanism. A Learned two-layer convolutional block processes concatenated magnitude and phase:
\begin{align}
    \mathbf{I}_\ell &= [\mathbf{A}_\ell; \mathbf{\Phi}_\ell] \label{eq:gate_input} \\
    \mathbf{G}_\ell &= \sigma(\Psi_{\text{gate}}(\mathbf{I}_\ell)) \in \mathbb{R}^{B \times C_{\text{bn}} \times H_\ell \times W_\ell} \label{eq:method-decoder-gate_output}
\end{align}

The gate is meant to identify spatial locations where phase encodes significant structure (edges, singularities) and permits bottleneck features to flow through via high values; attenuate signal in homogeneous regions via low values. Applied as multiplicative mask:
\begin{equation}
    \hat{U}_\ell = \tilde{U}_\ell \odot G_\ell, \label{eq:method-decoder-modulation}
\end{equation}
where $\odot$ denotes the element-wise product, $\tilde{U}_\ell$ denotes the bottleneck representation $Z$ upsampled bilinearly to spatial resolution $H_\ell \times W_\ell$ at the coarsest decoder level, or the output $U_{\ell+1}$ of the previous decoder stage at finer levels.

\paragraph{Fusion Network (Multi-Source Integration).} Three-source fusion combines gated bottleneck, magnitude, and phase:
\begin{align}
    \mathbf{I}_{fuse}^{(\ell)} &= [\hat{U}_\ell;A_\ell;\Phi_\ell] \label{eq:method-decoder-fusion_in} \\
    U_\ell &= \Psi_{fuse}(\mathbf{I}_{fuse}^{(\ell)}) \in \mathbb{R}^{B \times C_{bn} \times H_\ell \times W_\ell} \label{eq:method-decoder-fusion_out}
\end{align}

The fusion network is a two-layer convolutional block (Conv2d $\rightarrow$ BatchNorm $\rightarrow$ ReLU $\rightarrow$ Conv2d $\rightarrow$ BatchNorm $\rightarrow$ ReLU) learning optimal combinations of coarse context, texture details (magnitude), and geometric structure (phase).

\begin{figure}[ht]
  \vskip 0.2in
  \begin{center}
    \centerline{\includegraphics[width=\columnwidth]{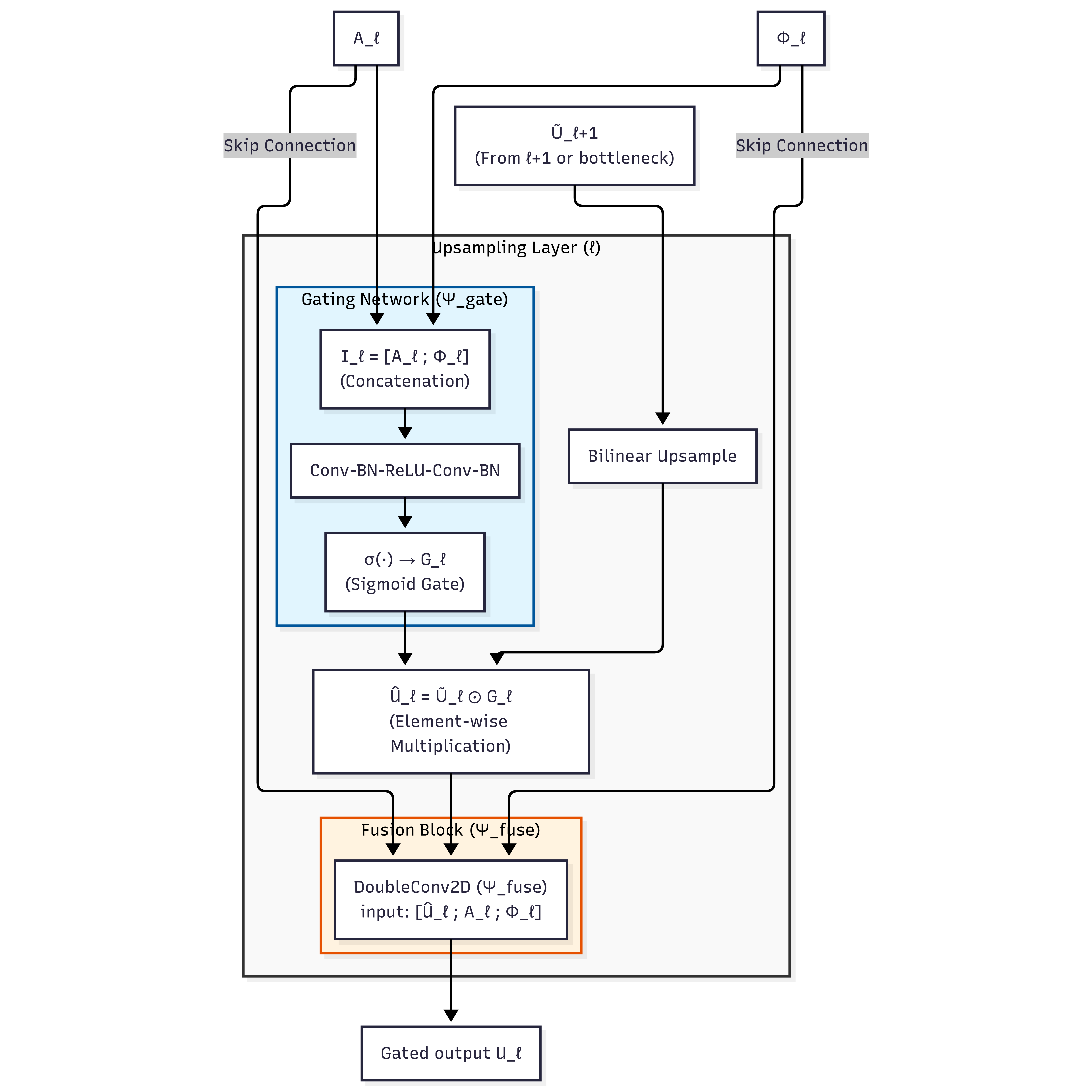}}
    \caption{
      Polar phase-aware learned gating mechanism --- Showing two components: Gating and Fusion. $\tilde{U}_\ell$: upsampled bottleneck features (bilinear interpolation from $Z$ at the coarsest level, or from the previous decoder output $U_{\ell+1}$ at finer levels).
    }
    \label{fig:method-decoder-fig}
  \end{center}
\end{figure}

In the decoder, the phase serves dual roles (see figure \ref{fig:method-decoder-fig}): learned attention mechanism in gating (where to focus) and raw input signal in fusion (what to focus on). This enables phase-driven geometric structure exploitation while preserving task-specific flexibility.

\paragraph{Output:} Pixel-level predictions after final upsampling and projection to task output space (denoising: 1 channel; segmentation: class probabilities).

\subsection{Mathematical Interpretability and Task-Specific Learning}
\label{subsec:method-interpret}
Mathematical interpretability in this work refers specifically to the encoder: the fixed complex Morlet filters $\psi^\theta_j$ have known frequency and orientation selectivity, and each scattering coefficient $S^{(\theta)}_j(u)$ carries a physically interpretable meaning as energy at scale $j$, orientation $\theta$, and spatial location $u$. The decoder modules $\Psi_\text{gate}$ and $\Psi_\text{fuse}$ are learned and not analytically interpretable; all interpretability claims in this paper are scoped to the encoder representation.

Analytically inverting the scattering transform via gradient descent phase retrieval requires $\sim$200 convergent iterations \cite{waldspurger2017wavelet}, making full signal recovery prohibitively expensive. We instead employ a task-specific learned decoder that operates directly on scattering coefficients. This trades analytical invertibility for task-optimized reconstruction: rather than recover the full input signal, we learn to directly predict task-relevant outputs. Critically, the fixed scattering encoder retains Lipschitz stability to input deformations \cite{bruna2013invariant}, providing theoretical robustness independent of decoder learning.

The decoder comprises three stages: (i) phase-informed gating that identifies regions where phase encodes task-relevant structure, (ii) learned fusion layers that integrate multiscale cues, and (iii) refinement layers that progressively refine predictions. Phase information guides this task-specific decoding, enabling efficient reconstruction optimized for the target task rather than generic signal recovery.

\section{Experiments}
\label{sec:exp}

We validate our phase-aware scattering wavelet-scattering-based encoder-decoder model on the standard image denoising benchmark BSD68 with Gaussian noise level $\sigma=25$. Our method \textit{Polar phase-aware with learned gating} \textit{(hereafter M6)} achieves $27.41 \pm 0.13$~dB PSNR. We compare against two strong baselines: \textit{DnCNN}, a learned residual CNN achieving $29.27 \pm 0.01$~dB PSNR, and a fully learned wavelet baseline \textit{DSWN} \cite{liu2020densely} achieving $29.61 \pm 0.04$ dB PSNR. \textit{Phase-aware} underperforms both baselines: $1.86$~dB gap relative to \textit{DnCNN} and $2.20$~dB gap relative to \textit{DSWN}. For broader context, BM3D achieves $28.57$~dB and SwinIR-M achieves $29.50$~dB; the $\sim 2$~dB gap is consistent across all fully learned methods. This performance gap reflects an explicit architectural trade-off: we sacrifice reconstruction quality for mathematical interpretability, Lipschitz stability guarantees derived from cascaded wavelet operations, and fixed multi-scale decomposition structure.

\begin{table}[t] 
  \caption{Baseline Performance Comparison on BSD68 ($\sigma=25$). \textit{Polar phase-aware with learned gating} achieves $27.41$~dB, underperforming learned methods by $1.86$~dB: \textit{DnCNN} \cite{7839189}, \textit{BM3D} \cite{dabov2007image}, \textit{SwinIR} \cite{9607618}. \textit{DSWN} \cite{liu2020densely} is a fully learned architecture using DWT/IDWT with dense residual blocks; both it and DnCNN are learned methods. The consistent $\sim 2$~dB gap across all learned baselines quantifies the cost of a fixed, mathematically interpretable encoder.} 
  \label{tab:exp-baseline}
  \begin{center}
    \begin{small}
      \begin{sc}
        \begin{tabular}{@{}lcc@{}}
          \toprule
          Method & PSNR (dB) & SSIM \\
          \midrule
          DnCNN (pretrained) & $29.273 \pm 0.013$ & $0.842 \pm 0.001$ \\
          BM3D (reported) & $28.57^\dagger$ & $0.801^\dagger$ \\
          SwinIR (reported) & $29.50^\dagger$ & $0.840^\dagger$ \\
          DSWN & $29.614 \pm 0.036$ & $0.855 \pm 0.002$ \\
          \midrule
          M6 (Ours) & $27.414 \pm 0.134$ & $0.797 \pm 0.002$ \\
          \bottomrule
          \multicolumn{3}{p{0.4\textwidth}}{\scriptsize $^\dagger$ indicates values reported from original papers without confidence intervals.}
        \end{tabular}
      \end{sc}
    \end{small}
  \end{center}
  \vskip -0.1in 
\end{table}

\subsection{Architectural Ablations: Stride-1 Equivariance and Phase Preservation}
\label{sec:exp-ablation}

To systematically isolate architectural contributions, we progressively build the method from baseline invariant scattering to the full phase-aware design. We begin with \textit{Invariant scattering with global smoothing}, baseline invariant scattering with standard global smoothing applied uniformly across all scales (operator $2^J$). This configuration achieves $24.19$~dB PSNR. By adopting stride-1 processing (avoiding subsampling) combined with per-scale scaled smoothing (each scale receives smoothing proportional to $2^j$ rather than $2^J$), we transition to \textit{Stride-1 equivariant with per-scale smoothing} and achieve $26.36$~dB PSNR—a gain of $+2.17$~dB. This substantial improvement validates the core hypothesis: spatial equivariance, not global invariance, is essential for dense prediction tasks. Critically, scattering's Lipschitz stability property—derived from cascaded modulus operations in the wavelet pyramid—is fully preserved under stride-1 processing; only the global invariance property is sacrificed.

Next, we upgrade skip connections from modulus-only representations to full complex-valued representations. Transitioning from \textit{Stride-1 equivariant with per-scale smoothing} to \textit{Complex-valued skip connections, no gating}, we achieve $27.39$~dB PSNR, an additional $+1.03$~dB improvement. This gain quantifies the value of preserving phase information in the encoder-decoder pathway. We further test polar representation, which decomposes complex values into magnitude and phase components and processes them separately, achieving $27.41$~dB PSNR—a marginal additional $+0.02$~dB improvement over raw complex representation. Polar form primarily enhances optimization stability rather than final performance; the critical contribution is the presence of phase information itself.

In synthesis, architectural contributions combine additively: $+2.17$~dB from stride-1 equivariance and $+1.03$~dB from phase preservation yield a total $+3.20$~dB improvement over baseline invariant scattering. This substantially reduces the performance gap: \textit{Invariant scattering with global smoothing} underperforms \textit{DnCNN} by $5.08$~dB, while \textit{Polar phase-aware with learned gating} underperforms by $1.86$~dB—a 63\% reduction in the gap through architectural refinement alone.

\begin{table}[t] 
  \caption{Architectural Ablation Progression. Each row isolates one architectural choice: (i) Baseline performing standard smoothing and modulus skips, (ii) Stride-1 with per-scale smoothing (modulus skips) provides the largest contribution ($+2.17$~dB), (iii) Complex-valued skip connections (scaled smooth) add $+1.03$~dB, and (iv) Polar representation (scaled smooth) adds marginal $+0.02$~dB. Total: $+3.20$~dB from \textit{Invariant scattering with global smoothing} to \textit{Polar phase-aware with learned gating}.} 
  \label{tab:exp-ablation}
  \begin{center}
    \begin{small}
      \begin{sc}
        \begin{tabular}{lcccc}
          \toprule
          Key Finding & PSNR & SSIM & $\Delta$ PSNR \\
          \midrule
          (i) & 24.192 & 0.714 & —  \\
          (ii) & 26.359 & 0.781 & +2.17  \\
          (iii) & 27.393 & 0.796 & +1.03 \\
          (iv) -- ours & 27.414 & 0.797 & +0.02 \\
          \bottomrule
        \end{tabular}
      \end{sc}
    \end{small}
  \end{center}
  \vskip -0.1in 
\end{table}

\subsection{Spatial Shuffling Analysis: Quantifying Phase Information Value}
\label{sec:exp-shuffle}

\begin{figure*}[t]
  \vskip 0.2in
  \begin{center}
    \begin{subfigure}[b]{0.32\textwidth}
      \centering
      \includegraphics[width=\linewidth]{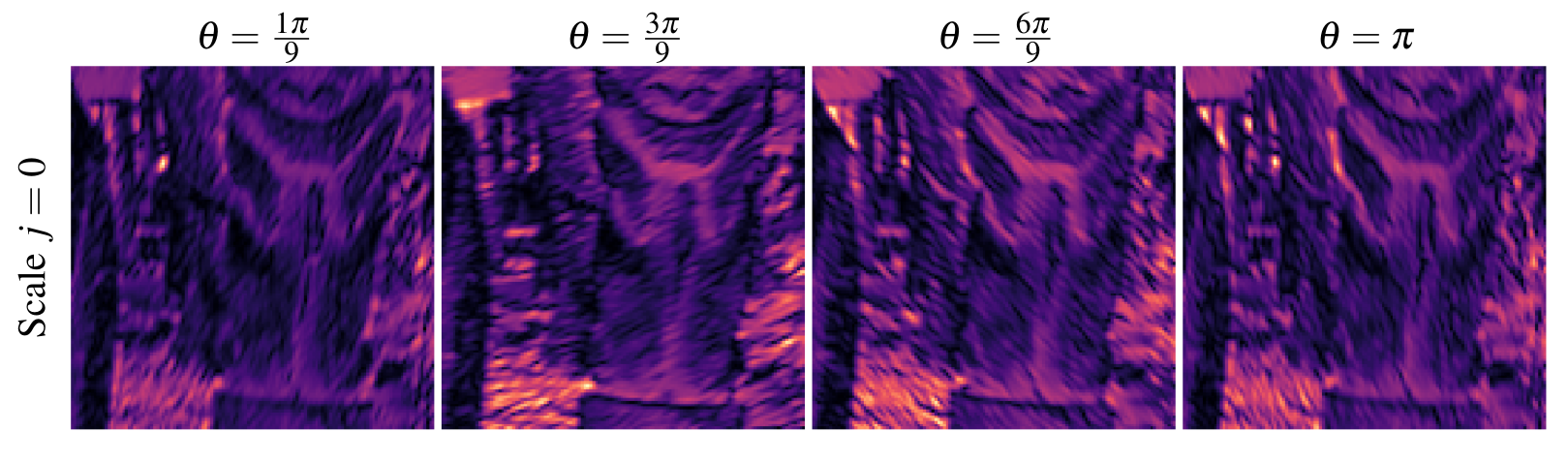}
      \caption{Magnitude maps $A^\theta_0$: Low-frequency, diffuse directional energy.}
      \label{fig:subfig-magnitude}
    \end{subfigure}
    \hfill
    \begin{subfigure}[b]{0.32\textwidth}
      \centering
      \includegraphics[width=\linewidth]{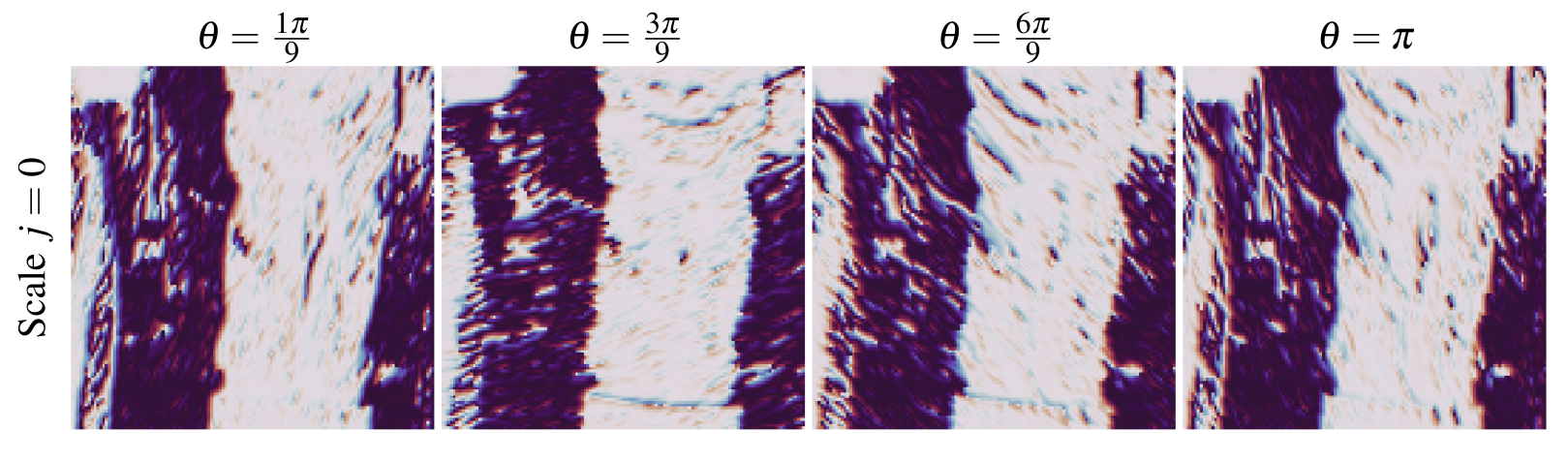}
      \caption{Phase maps $\Phi^\theta_0$: High-frequency spatial transitions.}
      \label{fig:subfig-phase}
    \end{subfigure}
    \hfill
    \begin{subfigure}[b]{0.32\textwidth}
      \centering
      \includegraphics[width=\linewidth]{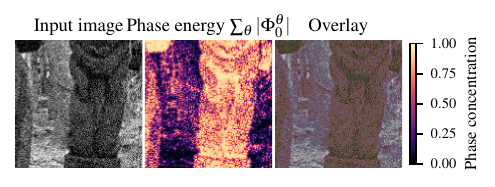}
      \caption{Phase energy overlay: Sharp geometric structures and edges.}
      \label{fig:subfig-overlay}
    \end{subfigure}

    \caption{
      Visualizing the representational split at scale $j=0$. (a) Magnitude maps $A^\theta_0$ show low-frequency, diffuse directional energy. (b) Phase maps $\Phi^\theta_0$ capture high-frequency spatial transitions. (c) Phase energy overlay confirms that phase explicitly encodes the sharp geometric structures and edges, explaining the severe -1.26 dB penalty when phase is spatially shuffled (Table \ref{tab:exp-shuffle}).
    }
    \label{fig:overlays-mags-fig}
  \end{center}
\end{figure*}

Phase information could encode either location-dependent structure (spatial correspondence) or texture statistics independent of location. To isolate these contributions, we perform a spatial shuffling ablation: we randomize the spatial indices of phase coefficients in skip connections while preserving phase channel statistics, and separately randomize amplitude indices. Polar phase-aware with learned gating baseline achieves $27.73$~dB PSNR. Phase-only shuffling ($\Phi$) drops performance to $26.47$~dB PSNR ($-1.26$~dB penalty). Shuffling all skip channels simultaneously drops to 26.30 dB PSNR ($-1.43$~dB penalty), confirming that the dominant loss comes from disrupting phase spatial correspondence. Amplitude shuffling drops to $27.48$~dB PSNR ($-0.25$~dB penalty). Since texture statistics are preserved in both shuffling operations, these penalties directly quantify location-dependent information content. The ratio is striking: phase information is $5.14\times$ more important than amplitude for reconstruction quality (see figure~\ref{fig:overlays-mags-fig}).

This finding clarifies the mechanism: phase encodes spatially-localized, coherent structures; texture alone is insufficient. The decoder learns to leverage this phase information through per-scale gating networks. When spatial phase-pattern correspondence is broken via shuffling, the decoder loses the ability to perform spatial modulation, resulting in severe reconstruction degradation. When amplitude is shuffled (texture disrupted), the impact is minimal because magnitude primarily encodes energy and frequency content, not location-dependent structure. This experiment provides direct evidence that phase information's value derives from spatial correspondence rather than texture statistics alone.

\begin{table}[t] 
  \caption{Spatial Shuffling Ablation Isolating Phase Contribution. Randomizing phase spatial indices (preserving statistics) costs $-1.26$~dB. Randomizing amplitude costs only $-0.25$~dB. Phase is $5.14\times$ more critical than amplitude, directly quantifying that phase encodes location-dependent information essential for reconstruction.} 
  \label{tab:exp-shuffle}
  \begin{center}
    \setlength{\tabcolsep}{2.5pt}
    \begin{small}
      \begin{sc}
        \begin{tabular}{lcccc}
          \toprule
          Shuf. & PSNR & SSIM & $\Delta$ PSNR & $\Delta$ SSIM \\
          \midrule
          $\emptyset$ & $27.730 \pm 0.090$ & $0.7997 \pm 0.0023$ & — & — \\
          All & $26.295 \pm 0.111$ & $0.7647 \pm 0.0049$ & $-1.435$ & $-0.0350$ \\
          $\Phi$ & $26.466 \pm 0.190$ & $0.7666 \pm 0.0050$ & $-1.264$ & $-0.0331$\\
          $A$ & $27.484 \pm 0.125$ & $0.7945 \pm 0.0034$ & $-0.246$ & $-0.0052$ \\
          \% & — & — & 5.14$\times$ & 6.37$\times$ \\
          \bottomrule
        \end{tabular}
      \end{sc}
    \end{small}
  \end{center}
  \vskip -0.1in 
\end{table}

\subsection{Sensitivity Studies}
\label{sec:exp-sensitivity}

\paragraph{Scattering Depth ($J$):} A central design choice is the depth J, which controls the number of cascaded wavelet scales. We find non-monotonic behavior: $J=2$ achieves $27.88$~dB PSNR with standard deviation $0.027$~dB (most stable, see Table~\ref{tab:exp-sensit-depth}). Our selected $J=3$ achieves $27.39$~dB PSNR with standard deviation $0.099$~dB. $J=4$ achieves $26.36$~dB PSNR. The gap between $J=2$ and $J=3$ is $+0.49$~dB, with $J=2$ being $3.7\times$ more stable (variance ratio $0.099/0.027 = 3.7$). We attribute this finding to task-specific structure: deeper scattering applies increasingly large smoothing windows. We selected $J=3$ for our primary method to align with standard scattering implementations (Kymatio library default), ensuring reproducibility and enabling direct comparison with the scattering literature. Practitioners requiring maximum performance should empirically tune $J$ based on patch size and detail requirements; $J=2$ is likely optimal for BSD68.

\paragraph{Wavelet Slant ($s$):} The slant parameter $s$ controls orientation selectivity. $s=0.0$ (horizontal-vertical orientations only) causes catastrophic performance collapse to $24.86$~dB PSNR, a $-2.54$~dB penalty versus our default (see Table~\ref{tab:exp-sensit-slant}). This is expected: images contain diagonal edges and structures; excluding diagonal wavelets removes critical information. Our default $s=0.5$ (balanced orientation coverage) achieves $27.40$~dB PSNR and is optimal. $s=1.0$ (higher directional emphasis) achieves $27.04$~dB PSNR ($-0.36$~dB), suboptimal but more stable.

\paragraph{Decoder Base Channels:} Base decoder channels control the width of learned transformation layers (see Table~\ref{tab:exp-sensit-depth}). Increasing from $32$ to $64$ channels improves PSNR by $+0.235$~dB (0.85\% improvement). Increasing from $64$ to $128$ channels provides only $+0.002$~dB improvement—negligible. The architecture exhibits capacity saturation at 64 channels. This is significant: the method is information-limited, not parameter-limited. Performance ceiling is determined by the quality of phase and multi-scale structure in skip connections, not model size.

\paragraph{Gating Network Hidden Channels:} Gating network hidden channel count (tested $16$ to $64$ channels) shows negligible impact: all configurations cluster within $\pm0.05$~dB of baseline (see Table~\ref{tab:gating}). This result is consistent with our finding that learned gates converge to spatially-constant scaling (Section~\ref{sec:analysis}). Gate capacity is under-utilized; spatial attention is not learned.

\subsection{Perturbation and stability analysis}
\label{sec:exp-perturbation}
\begin{figure}[ht]
  \vskip 0.2in
  \begin{center}
    \centerline{\includegraphics[width=\columnwidth]{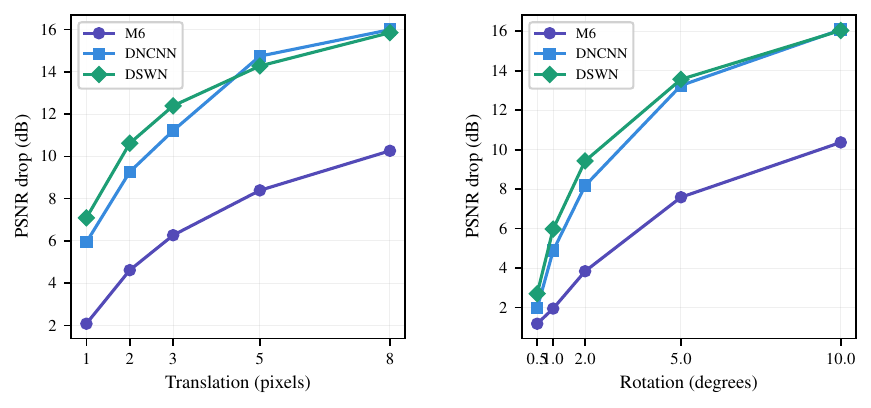}}
    \caption{
    \textbf{Perturbation robustness:} PSNR drop from unperturbed baseline under increasing translation (left) and rotation (right) for M6 (Ours), DnCNN, and DSWN on BSD68 ($\sigma = 25$). Perturbations are applied to the noisy input; all methods receive identically perturbed inputs. M6 (Ours) degrades more slowly than DnCNN and DSWN, consistent with the Lipschitz stability guarantee of the fixed scattering encoder.
    }
    \label{fig:exp-perturbation-perturbation-fig}
  \end{center}
\end{figure}

\begin{figure}[ht]
  \vskip 0.2in
  \begin{center}
    \centerline{\includegraphics[width=\columnwidth]{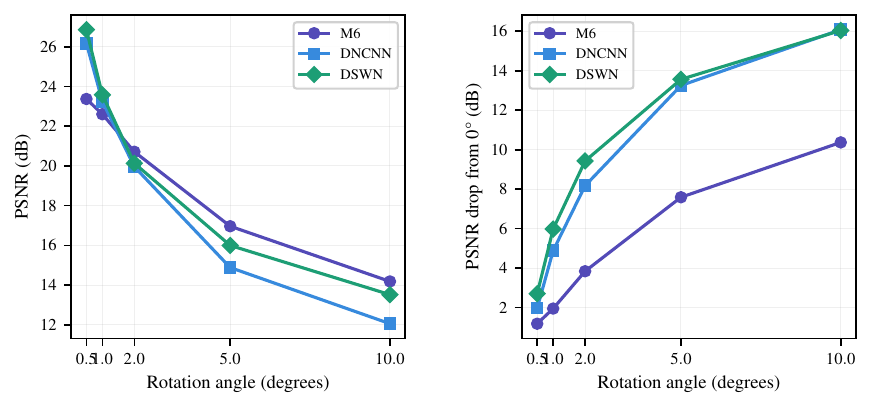}}
    \caption{
    \textbf{Stability comparison.} Absolute PSNR (left) and PSNR drop from unperturbed baseline (right) as a function of rotation angle. M6 (Ours) starts lower in absolute PSNR but degrades more slowly: at $10\circ$, M6 (Ours) drops $10.4$~dB vs. $16.1$~dB for DnCNN ($1.55 \times$ more robust). At large rotations the stability advantage partially compensates the clean-image performance gap.
    }
    \label{fig:exp-perturbation-stability-fig}
  \end{center}
\end{figure}

Table \ref{tab:analysis-perturbation} and Figures \ref{fig:exp-perturbation-perturbation-fig}–\ref{fig:exp-perturbation-stability-fig} present robustness to geometric perturbations on BSD68 ($\sigma = 25$); M6 (Ours) is $1.55 \times$ more robust than DnCNN under rotation, providing empirical evidence of the scattering encoder's Lipschitz stability.

\begin{table}[t] 
  \caption\textbf{{Robustness to geometric perturbations on BSD68 ($\sigma=25$).} Perturbations are applied to the noisy input; absolute PSNR drops are large because noise patterns do not transform coherently under rotation. The relative comparison is valid as all methods receive identically perturbed inputs. M6 (Ours) is $1.55 \times$ more robust than \textit{DnCNN} under rotation, providing empirical evidence of the scattering encoder's Lipschitz stability.} 
  \label{tab:analysis-perturbation}
  \begin{center}
    \begin{small}
      \begin{sc}
        \begin{tabular}{@{}lcccc@{}}
          \toprule
          Perturbation & M6 (Ours) & DnCNN & DSWN \\
          \midrule
          Rotation $5^\circ$ & $-7.6$ & $-13.3$ & $-13.6$ \\
          Rotation $10^\circ$ & $-10.4$ & $-16.1$ & $-16.0$  \\
          Translation $5$~px & $-8.4$ & $-14.7$ & $-14.3$ \\
          Translation $8$~px & $-10.3$ & $-16.0$ & $-15.9$ \\
          \bottomrule
        \end{tabular}
      \end{sc}
    \end{small}
  \end{center}
  \vskip -0.1in 
\end{table}

\subsection{Data Efficiency}
\label{sec:exp-efficiency}
We evaluate robustness to limited training data by assessing \textit{Polar phase-aware with learned gating} and \textit{DnCNN} across data fractions: 10\%, 25\%, 50\%, and 100\% of BSD400 training set (Table~\ref{tab:prel-seg}). \textit{DnCNN} is remarkably robust: at 10\% data (approximately $40$ training images and $\sim$100s augmented patches), \textit{DnCNN} achieves $28.77 \pm 0.24$~dB PSNR, degrading only 1.71\% from full-data performance ($29.27$~dB). \textit{Polar phase-aware with learned gating} is fragile by comparison: at 10\% data, phase-aware achieves only $25.96 \pm 0.07$~dB PSNR, degrading 5.29\% from full-data performance ($27.41$~dB).

This creates a significant performance gap at low data: at 10\%, the gap widens to $-2.81$~dB (\textit{Polar phase-aware with learned gating} vs \textit{DnCNN}), compared to the full-data gap of $-1.86$~dB. Computing the robustness ratio: \textit{DnCNN} is $3.1\times$ more robust to data scarcity than \textit{Polar phase-aware with learned gating} (5.29\% / 1.71\% $\approx$ 3.1). This difference reflects architectural fundamentals. Phase-aware includes approximately 100K learnable parameters distributed across bottleneck, gating, and fusion networks. These components must be supervised to learn task-specific mappings; with 10\% data, they severely underfit. In contrast, \textit{DnCNN}'s convolutional layers benefit from strong inductive biases (local spatial patterns are universal across images) that generalize well with limited supervision. This limitation defines applicability bounds: \textit{Polar phase-aware with learned gating} is suitable for well-resourced settings with $N > 100$ training images. For severely data-limited settings ($N < 50$), alternative methods are preferable.
Consequently, the method is not well-suited for typical medical imaging domains, where annotated datasets often number fewer than $100$ samples.

\subsection{Preliminary Generalization Study (Training-Only, No Validation)}
\label{sec:exp-gen}
\textbf{All results in this section are training-set performance only. No validation or test evaluation is reported. This section constitutes a proof-of-concept extensibility study; we make no claim of validated segmentation performance.}

To validate whether architectural insights generalize beyond image denoising, we conducted preliminary segmentation studies on ISIC 2018 (International Skin Imaging Collaboration dataset) for binary skin lesion segmentation. Using configuration $J=2$ (finer detail than denoising) and reduced channels (8 base, 4 gate hidden), we trained for 30 epochs on the ISIC training split (approximately 1800 grayscale images). On training data, the architecture achieves consistent Dice coefficients of $0.886$--$0.887$ across multiple experimental configurations.

\paragraph{Architecture robustness across gate activation functions:} We tested ReLU, Sigmoid, and LeakyReLU gate activations. Sigmoid and LeakyReLU marginally outperformed ReLU by approximately $+0.001$ Dice ($0.8858$ for ReLU vs $0.8867$ for Sigmoid, $0.8871$ for LeakyReLU). This suggests gate activation function choice has negligible impact on segmentation performance.

\paragraph{Phase regularization provides no benefit on segmentation:} Phase regularization losses (Phase Total Variation and Phase Alignment) provided no performance improvement on segmentation ($\Delta \leq -0.0003$ Dice). This extends the denoising null result across tasks: explicit phase constraints are ineffective on both image reconstruction (denoising) and discriminative (segmentation) tasks. Implicit phase coherence learning suffices for both. (see Appendix \ref{app:prel_seg})

\paragraph{Critical caveat—validation results are incomplete.} This preliminary segmentation study has a fundamental limitation: training-only Dice of $0.886$--$0.887$ does not constitute proof of generalization. Validation performance is required. We report training results as proof-of-concept that the architecture generalizes to medical image segmentation; we do not claim validated segmentation capability. Full cross-validation with extended training (200+ epochs), proper hyperparameter tuning, and comparison to established segmentation baselines (\textit{nnU-Net}, \textit{Swin-UNet}) remain ongoing.

\section{Analysis}
\label{sec:analysis}

\begin{figure*}[t]
  \vskip 0.2in
  \begin{center}
    \centerline{\includegraphics[width=\textwidth]{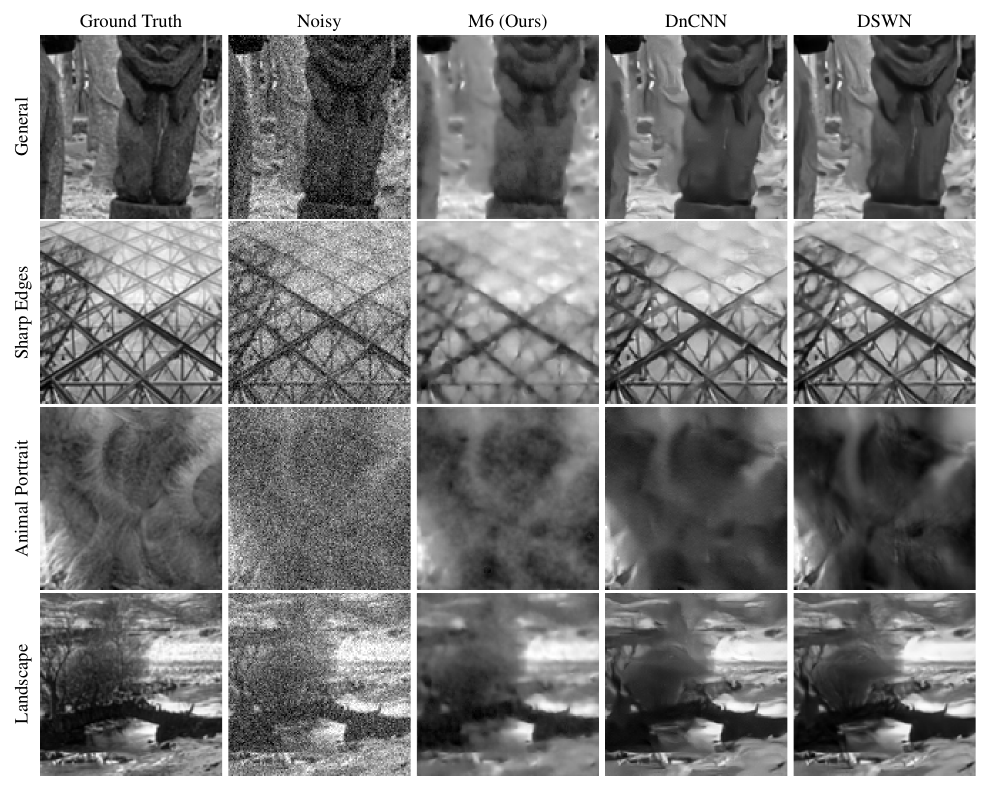}}
    \caption{
      \textbf{Qualitative denoising results on BSD68 ($\sigma=25$).} While DnCNN (learned) achieves higher absolute PSNR by aggressively smoothing, M6 (Ours) leverages the phase-aware scattering encoder to preserve structural integrity without relying on a black-box feature extractor.
    }
    \label{fig:qualitative-fig}
  \end{center}
\end{figure*}

The core contribution of this work is demonstrating that Lipschitz stability and spatial invariance—traditionally treated as inseparable properties in classification-based scattering networks—can be decoupled for dense prediction tasks. In prior scattering architectures, cascaded modulus operations provide both stability and invariance simultaneously. However, for dense prediction, global invariance is prohibitive as it discards spatial structure, whereas stability remains valuable for providing robustness and mathematical interpretability. Our approach preserves cascaded modulus operations to maintain Lipschitz stability while breaking global invariance through stride-1 processing.

Quantitatively, architectural contributions aggregate to $+3.20$~dB improvement over baseline invariant scattering. Stride-1 equivariance contributes $+2.17$~dB, phase preservation contributes $+1.03$~dB, and polar representation contributes $+0.02$~dB. This composition reduces the performance gap from $5.08$~dB (\textit{Stride-1 equivariant with per-scale smoothing} vs. \textit{DnCNN}) to $1.86$~dB (\textit{Polar phase-aware with learned gating} vs. \textit{DnCNN})—a 63\% reduction. The dominance of stride-1 contribution validates the hypothesis that spatial equivariance is essential for dense prediction. The substantial $+1.03$~dB phase contribution confirms that phase information encodes location-dependent coherent structures critical for reconstruction tasks.

We evaluated M6 (Ours), DnCNN, and DSWN under input rotations ($5^\circ$ - $10^\circ$) and translations (1–8 px) on BSD68 ($\sigma = 25$) as depicted in Table~\ref{tab:analysis-perturbation}. M6 (Ours) is $1.55 \times$ more robust than DnCNN at rotation (10° vs. $1.61$ dB drop). At large perturbations ($\geq 5^\circ$), the stability advantage narrows and can invert the clean-image performance gap—demonstrating the practical value of the scattering encoder's Lipschitz stability.

\subsection{What Didn't Work}
\label{subsec:analysis-didnt}
Two design choices underperformed their theoretical motivation, yielding valuable insights into what mechanisms genuinely matter versus what theory predicts:

\paragraph{Finding 1:} Gating Networks Converge to Learned Constant Scaling, Not Spatial Attention.
We hypothesized that learned gating networks would perform phase-guided spatial attention.
Instead, gates converged to spatially-constant per-scale masks (spatial variance $< 0.02$, far below the $0.1$ threshold for meaningful spatial variation).
Gating contributes only $+0.02$~dB to overall performance---negligible compared to phase preservation ($+1.03$~dB).
This reveals a fundamental constraint: spatial attention is difficult to learn from high-dimensional phase information; networks preferentially discover simpler solutions (constant per-scale gains).
The mechanistic insight is that phase presence itself drives performance gains, not the sophistication of gating mechanisms.

\paragraph{Finding 2:} Explicit Phase Regularization Conflicts with Task Objectives.
We tested Phase Total Variation (enforcing smooth phases) and Phase Alignment (enforcing cross-scale consistency), both motivated by scattering theory.
Both underperformed baseline: $-0.023$~dB in denoising and $-0.0003$ Dice in segmentation.
The mechanism is clear: edges possess sharp phase gradients that encode polarity and orientation; enforcing smoothness destroys this information.
Implicit phase coherence learning via reconstruction loss suffices.
This demonstrates that hand-tuned mathematical constraints, while theoretically motivated, can inadvertently constrain task-specific learning. These results clarify which mechanisms actually drive performance versus which are theoretically appealing but practically inert.

\subsection{Task-Dependent Effects}
\label{subsec:analysis-task}
Preliminary segmentation studies reveal task-dependent patterns in phase contribution.
In denoising, phase preservation provides $+1.03$~dB PSNR (statistically significant); in segmentation, phase regularization provides no measurable benefit ($\Delta \leq -0.0003$ Dice), extending the denoising null result.
This suggests phase information's contribution varies by task objectives.
One hypothesis: denoising requires coherent structure recovery (edges, singularities), where phase encodes spatial correspondence; segmentation may rely more heavily on magnitude information (frequency decomposition).
However, this hypothesis remains speculative and requires careful validation through full ISIC cross-validation (Section~6 Key Research Directions), baseline comparisons to \textit{nnU-Net} and \textit{Swin-UNet}, and extended training protocols.
What we can confidently establish is that implicit phase learning via task loss suffices for both denoising and segmentation.

\subsection{Prior Wavelet-Based Methods}
\label{subsec:analysis-wavelet}
Wavelet integration into deep learning has historically followed two distinct paradigms:

\paragraph{Approach 1:} Learned Wavelets.
Methods such as \textit{MWCNN} and \textit{SDWNet} learn wavelet filters end-to-end, optimizing task performance at the cost of interpretability.
Learned wavelet filters become black-box parameters.
Empirical performance is typically strong ($29+$~dB).
DSWN \cite{liu2020densely} similarly learns DWT/IDWT filters with dense residual blocks, achieving $29.61$~dB.

\paragraph{Approach 2:} Fixed Wavelets.
Classical scattering networks maintain interpretability and mathematical rigor through fixed, theoretically-grounded wavelet bases, sacrificing performance.

Our contribution bridges these approaches: we employ fixed, mathematically-grounded scattering wavelets (preserving mathematical interpretability and stability guarantees) while introducing architectural innovations---stride-1 equivariance and phase-aware skip connections---that enable spatial structure recovery for dense prediction.

Practitioner guidance follows naturally: Choose learned wavelets if maximum performance is the sole priority ($29+$~dB). DSWN, the fully learned wavelet method, achieves $29.61$~dB.
Choose \textit{Polar phase-aware with learned gating} if large data is available, and interpretability, provable stability, and mathematical transparency are requirements.
Medical imaging applications, typically constrained by annotation scarcity, remain better served by methods designed for low-data regimes.
The $1.86$~dB gap versus \textit{DnCNN} and $2.20$~dB gap versus DSWN represents the principled cost of mathematical interpretability---meaningful but justified for applications demanding it.

\section{Future Work}
\label{sec:future_work}
Data efficiency improvements through semi-supervised pre-training or self-supervised contrastive learning could reduce \textit{Polar phase-aware with learned gating}'s sample complexity.
Phase dropout (stochastic perturbation of phase to decorrelate learned patterns) offers a promising avenue for mitigating segmentation overfitting.

Task-specific depth tuning merits investigation to recover the $+0.49$~dB improvement ($J=2$ vs $J=3$).
Learnable depth schedules and adaptive scattering enabling dynamic multi-scale selection based on image content warrant further study.

Gating simplification---replacing learned gating networks with fixed per-scale gains---could reduce parameters and improve interpretability without sacrificing performance.

Full ISIC cross-validation (200+ epochs with proper early stopping), comprehensive hyperparameter tuning, and rigorous comparison to \textit{nnU-Net} and \textit{Swin-UNet} baselines are essential to clarify task-dependent phase effects in segmentation.

Systematic validation of the Lipschitz stability hypothesis through perturbation studies (small affine transformations, elastic deformations) would benchmark \textit{Polar phase-aware with learned gating}'s robustness against learned methods (\textit{DnCNN}, \textit{SwinIR}).

Multi-channel extension via reduced scattering depth ($J=2$) enables natural image segmentation.
Mixed-precision training merits exploration to reduce memory footprint and enable larger batch sizes.

\section{Conclusion}
\label{sec:conclusion}
We demonstrate that scattering's Lipschitz stability can be decoupled from spatial invariance for dense prediction.
Stride-1 equivariance ($+2.17$~dB) and phase-aware reconstruction ($+1.03$~dB) enable spatial structure recovery while maintaining theoretical robustness guarantees.
Comprehensive ablations, including honest null findings (gating, regularization), demonstrate scientific rigor.
Our work advances the understanding of what mechanisms matter in wavelet-based deep learning: spatial localization and phase information are essential; hand-engineered constraints can harm task-specific learning.

\bibliography{references}

@article{mallat2012group,
  title={Group invariant scattering},
  author={Mallat, St{\'e}phane},
  journal={Communications on Pure and Applied Mathematics},
  volume={65},
  number={10},
  pages={1331--1398},
  year={2012},
  publisher={Wiley Online Library}
}

@article{bruna2013invariant,
  title={Invariant scattering convolution networks},
  author={Bruna, Joan and Mallat, St{\'e}phane},
  journal={IEEE transactions on pattern analysis and machine intelligence},
  volume={35},
  number={8},
  pages={1872--1886},
  year={2013},
  publisher={IEEE}
}

@InProceedings{ronneberger2015,
author="Ronneberger, Olaf
and Fischer, Philipp
and Brox, Thomas",
editor="Navab, Nassir
and Hornegger, Joachim
and Wells, William M.
and Frangi, Alejandro F.",
title="U-Net: Convolutional Networks for Biomedical Image Segmentation",
booktitle="Medical Image Computing and Computer-Assisted Intervention -- MICCAI 2015",
year="2015",
publisher="Springer International Publishing",
address="Cham",
pages="234--241",
isbn="978-3-319-24574-4"
}

@article{CHEN2024103280,
title = {TransUNet: Rethinking the U-Net architecture design for medical image segmentation through the lens of transformers},
journal = {Medical Image Analysis},
volume = {97},
pages = {103280},
year = {2024},
issn = {1361-8415},
doi = {https://doi.org/10.1016/j.media.2024.103280},
url = {https://www.sciencedirect.com/science/article/pii/S1361841524002056},
author = {Jieneng Chen and Jieru Mei and Xianhang Li and Yongyi Lu and Qihang Yu and Qingyue Wei and Xiangde Luo and Yutong Xie and Ehsan Adeli and Yan Wang and Matthew P. Lungren and Shaoting Zhang and Lei Xing and Le Lu and Alan Yuille and Yuyin Zhou}
}

@inproceedings{cao2022swin,
  title={Swin-unet: Unet-like pure transformer for medical image segmentation},
  author={Cao, Hu and Wang, Yueyue and Chen, Joy and Jiang, Dongsheng and Zhang, Xiaopeng and Tian, Qi and Wang, Manning},
  booktitle={European conference on computer vision},
  pages={205--218},
  year={2022},
  organization={Springer}
}

@ARTICLE{7839189,
  author={Zhang, Kai and Zuo, Wangmeng and Chen, Yunjin and Meng, Deyu and Zhang, Lei},
  journal={IEEE Transactions on Image Processing}, 
  title={Beyond a Gaussian Denoiser: Residual Learning of Deep CNN for Image Denoising}, 
  year={2017},
  volume={26},
  number={7},
  pages={3142-3155},
  doi={10.1109/TIP.2017.2662206}
}

@INPROCEEDINGS{9607618,
  author={Liang, Jingyun and Cao, Jiezhang and Sun, Guolei and Zhang, Kai and Van Gool, Luc and Timofte, Radu},
  booktitle={2021 IEEE/CVF International Conference on Computer Vision Workshops (ICCVW)}, 
  title={SwinIR: Image Restoration Using Swin Transformer}, 
  year={2021},
  volume={},
  number={},
  pages={1833-1844},
  doi={10.1109/ICCVW54120.2021.00210}}

@inproceedings{liu2018multi,
  title={Multi-level wavelet-CNN for image restoration},
  author={Liu, Pengju and Zhang, Hongzhi and Zhang, Kai and Lin, Liang and Zuo, Wangmeng},
  booktitle={Proceedings of the IEEE conference on computer vision and pattern recognition workshops},
  pages={773--782},
  year={2018}
}

@inproceedings{zou2021sdwnet,
  title={Sdwnet: A straight dilated network with wavelet transformation for image deblurring},
  author={Zou, Wenbin and Jiang, Mingchao and Zhang, Yunchen and Chen, Liang and Lu, Zhiyong and Wu, Yi},
  booktitle={Proceedings of the IEEE/CVF international conference on computer vision},
  pages={1895--1904},
  year={2021}
}

@article{mallat2020phase,
  title={Phase harmonic correlations and convolutional neural networks},
  author={Mallat, St{\'e}phane and Zhang, Sixin and Rochette, Gaspar},
  journal={Information and Inference: A Journal of the IMA},
  volume={9},
  number={3},
  pages={721--747},
  year={2020},
  publisher={Oxford University Press}
}

@ARTICLE{Cole2021-oz,
  title    = "Analysis of deep complex-valued convolutional neural networks for
              {MRI} reconstruction and phase-focused applications",
  author   = "Cole, Elizabeth and Cheng, Joseph and Pauly, John and Vasanawala,
              Shreyas",
  journal  = "Magn Reson Med",
  volume   =  86,
  number   =  2,
  pages    = "1093--1109",
  month    =  mar,
  year     =  2021,
  address  = "United States",
  language = "en"
}

@article{DRAMSCH2021104643,
title = {Complex-valued neural networks for machine learning on non-stationary physical data},
journal = {Computers \& Geosciences},
volume = {146},
pages = {104643},
year = {2021},
issn = {0098-3004},
doi = {https://doi.org/10.1016/j.cageo.2020.104643},
url = {https://www.sciencedirect.com/science/article/pii/S0098300420306208},
author = {Jesper Sören Dramsch and Mikael Lüthje and Anders Nymark Christensen}
}

@ARTICLE{Chen2025,
  title   = "{Amplitude-Phase} Interaction Network for Speech Emotion
             Recognition",
  author={Haozhe Chen, Xiaojuan Zhang},
  journal = "APIN",
  year    =  2025
}

@article{GUO2025103201,
title = {APIN: Amplitude- and phase-aware interaction network for speech emotion recognition},
journal = {Speech Communication},
volume = {169},
pages = {103201},
year = {2025},
issn = {0167-6393},
doi = {https://doi.org/10.1016/j.specom.2025.103201},
url = {https://www.sciencedirect.com/science/article/pii/S0167639325000160},
author = {Lili Guo and Jie Li and Shifei Ding and Jianwu Dang}
}

@article{wangcomplex,
  title={Complex-valued Scattering Representations},
  author={Wang, Ke and Singhal, Utkarsh and Lustig, Michael and Yu, Stella X},
  year={2024},
}

@article{waldspurger2017wavelet,
  title={Wavelet transform modulus: phase retrieval and scattering},
  author={Waldspurger, Ir{\`e}ne},
  journal={Journ{\'e}es {\'e}quations aux d{\'e}riv{\'e}es partielles},
  pages={1--10},
  year={2017}
}

@article{nie2021neural,
  title={Neural network is heterogeneous: Phase matters more},
  author={Nie, Yuqi and Yuan, Hui},
  journal={arXiv preprint arXiv:2111.02014},
  year={2021}
}

@inproceedings{liu2020densely,
  title={Densely Self-Guided Wavelet Network for Image Denoising},
  author={Liu, Wei and Yan, Qiong and Zhao, Yuzhi},
  booktitle={Proceedings of the IEEE/CVF Conference on Computer Vision and Pattern Recognition Workshops},
  pages={432--433},
  year={2020}
}

@misc{andreux2022kymatio,
  author = {Andreux, M. and Angles, T. and Exarchakis, G. and Leonarduzzi, R. and Rochette, G. and Thiry, L. and Zarka, J. and Mallat, S. and Andén, J. and Belilovsky, E. and Bruna, J. and Lostanlen, V. and Chaudhary, M. and Hirn, M. J. and Oyallon, E. and Zhang, S. and Cella, C. and Eickenberg, M.},
  title = {Kymatio: Scattering Transforms in Python},
  year = {2022},
  url = {https://arxiv.org/abs/1812.11214},
  note = {MIT License}
}

@article{dabov2007image,
  title={Image denoising by sparse 3-D transform-domain collaborative filtering},
  author={Dabov, Kostadin and Foi, Alessandro and Katkovnik, Vladimir and Egiazarian, Karen},
  journal={IEEE Transactions on image processing},
  volume={16},
  number={8},
  pages={2080--2095},
  year={2007},
  publisher={IEEE}
}
\bibliographystyle{icml2026}

\newpage
\appendix
\onecolumn
\section{Scattering Encoder details}
\label{app:scat_enc}
\begin{figure}[ht]
  \vskip 0.2in
  \begin{center}
    \centerline{\includegraphics[width=\textwidth]{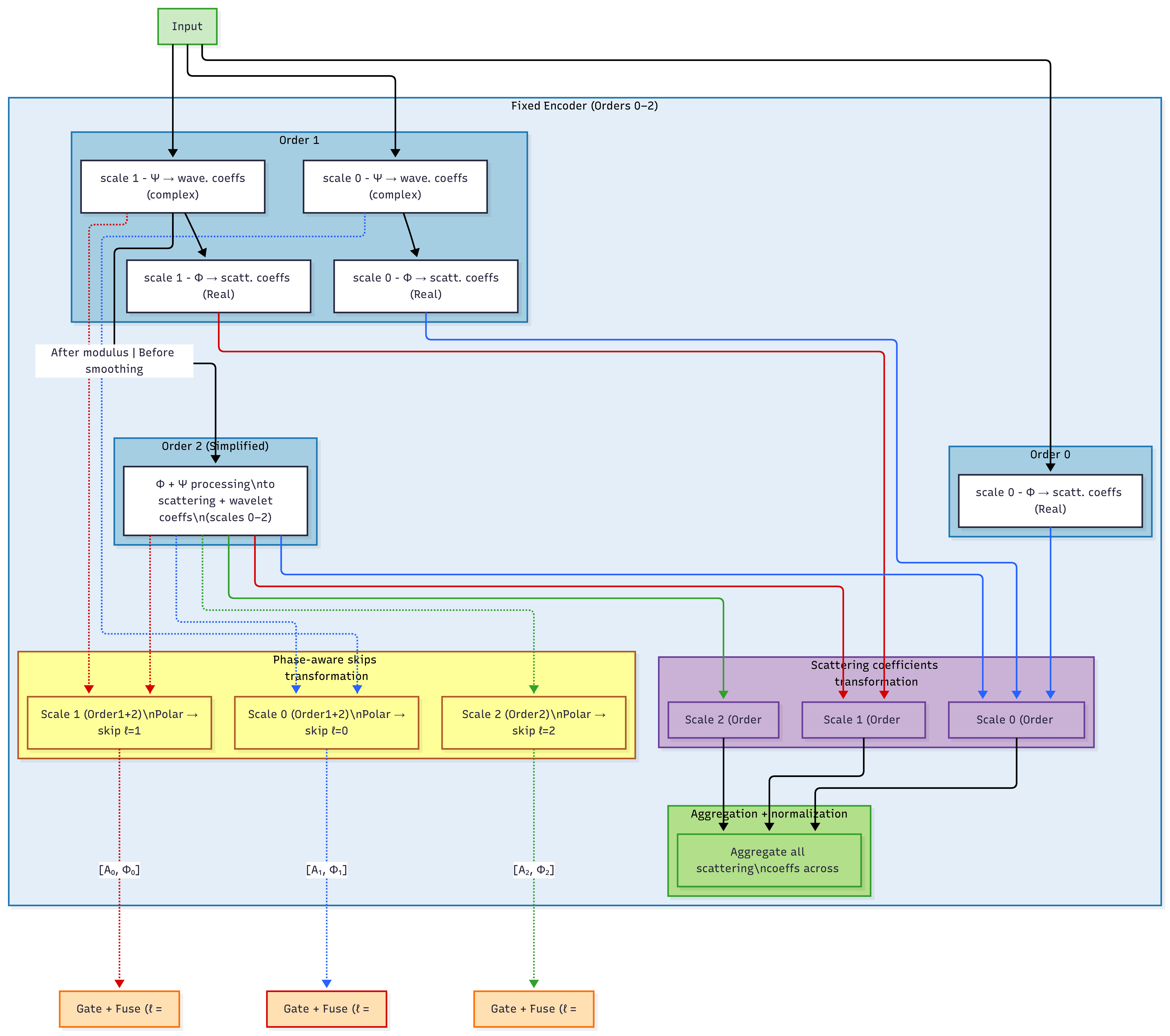}}
    \caption{
      \textbf{Detailed scattering encoder architecture.} Order-0 produces a global low-pass reference ($S_0$). Order-1 computes wavelet responses at each scale and orientation, then applies scale-dependent smoothing. Order-2 processes second-order paths (sibling paths). Pre-modulus complex coefficients at each scale are stored and converted to polar form $(A_j, \Phi_j)$ for use in phase-aware skip connections.
    }
    \label{fig:encoder-fig}
  \end{center}
\end{figure}

\section{Extended Sensitivity Analysis}
\label{app:exp-gen-phase}
The Phase Total Variation Loss encourages smooth phase fields by penalizing angular differences between adjacent pixels, while the Phase Alignment Loss enforces cross-scale consistency via downsampled fine-to-coarse phase comparisons:
\begin{align}
    \mathcal{L}_{\text{PTV}} &= \frac{1}{2} \sum_{d \in \{x, y\}} \frac{1}{N_d} \sum_{i,j} \rho(\nabla_d \Phi_{i,j}) \label{eq:ptv} \\
    \mathcal{L}_{\text{PAlign}} &= \frac{1}{(J-1)hw} \sum_{j=0}^{J-2} \sum_{x,y} \rho \left( \Phi_j^{\downarrow}, \Phi_{j+1} \right) \label{eq:palign}
\end{align}


\begin{table}[ht]
  \caption{Scattering Depth Sensitivity ($J$)}
  \label{tab:exp-sensit-depth}
  \begin{center}
    \begin{small}
      \begin{sc}
        \begin{tabular}{lccc}
          \toprule
          J & PSNR (dB) & Std Dev & Relative Stability \\
          \midrule
          2 & 27.88 & 0.027 & 1.0x (baseline) \\
          3 & 27.39 & 0.099 & 3.7x worse \\
          4 & 26.36 & --- & --- \\
          \bottomrule
        \end{tabular}
      \end{sc}
    \end{small}
  \end{center}
  \vskip -0.1in
\end{table}

\textbf{Finding:} J=2 most stable ($\sigma=0.027$~dB). Selected J=3 for reproducibility with Kymatio library default.

\begin{table}[ht]
  \caption{Wavelet Slant Sensitivity ($s$)}
  \label{tab:exp-sensit-slant}
  \begin{center}
    \begin{small}
      \begin{sc}
        \begin{tabular}{lccc}
          \toprule
          Slant (s) & PSNR (dB) & $\Delta$PSNR vs optimal & Notes \\
          \midrule
          0.0 & 24.86 & $-2.54$ & Catastrophic (no diagonal wavelets) \\
          0.5 & 27.40 & 0.0 & Optimal, balanced orientation \\
          1.0 & 27.04 & $-0.36$ & Suboptimal but more stable \\
          \bottomrule
        \end{tabular}
      \end{sc}
    \end{small}
  \end{center}
  \vskip -0.1in
\end{table}

\textbf{Rationale:} s=0.5 provides balanced coverage of horizontal, vertical, and diagonal edges.

\begin{table}[ht]
  \caption{Decoder Base Channels Sensitivity}
  \label{tab:exp-sensit-channels}
  \begin{center}
    \begin{small}
      \begin{sc}
        \begin{tabular}{lccc}
          \toprule
          Base Channels & PSNR Gain vs 32ch & Absolute PSNR \\
          \midrule
          32 & --- & 27.175 \\
          64 & +0.235~dB & 27.410 \\
          128 & +0.002~dB & 27.412 \\
          \bottomrule
        \end{tabular}
      \end{sc}
    \end{small}
  \end{center}
  \vskip -0.1in
\end{table}

\textbf{Interpretation:} Saturation at 64 channels. Model is information-limited (by phase/multi-scale quality), not parameter-limited.

\begin{table}[ht]
  \caption{Gating Network Hidden Channels Sensitivity}
  \label{tab:gating}
  \begin{center}
    \begin{small}
      \begin{sc}
        \begin{tabular}{lcc}
          \toprule
          Hidden Channels & PSNR & Variance \\
          \midrule
          16 & $27.406 \pm 0.04$~dB & Within $\pm0.05$~dB \\
          32 & $27.414 \pm 0.03$~dB & Within $\pm0.05$~dB \\
          64 & $27.410 \pm 0.05$~dB & Within $\pm0.05$~dB \\
          \bottomrule
        \end{tabular}
      \end{sc}
    \end{small}
  \end{center}
  \vskip -0.1in
\end{table}

\textbf{Finding:} Negligible impact. Gates learn spatially-constant scaling (variance $< 0.02$), not spatial attention.

\section{Preliminary Segmentation Extensibility Study (Training-Only).}
\label{app:prel_seg}
Table \ref{tab:exp-seg-tab} presents the segmentation ablation results on training data only. All conclusions drawn from this table are preliminary; no validation performance is reported.

\begin{table}[ht] 
  \caption{ISIC 2018 Segmentation Ablations: Loss Configuration and Gate Activation Impact. Phase regularization yields no measurable improvement, consistent with denoising findings. All configurations converge to Dice $\approx 0.886$--$0.887$ on training data.} 
  \label{tab:exp-seg-tab}
  \begin{center}
    \setlength{\tabcolsep}{2.5pt}
    \begin{small}
      \begin{sc}
        \begin{tabular}{lccccc}
          \toprule
          Loss Config & Gate & Epochs & Train Dice & Finding \\
          \midrule
          Dice + CE & ReLU & 11 & $0.8858 \pm 0.0015$ & Baseline \\
          Dice + CE + Skip & Sigmoid & 30 & $0.8867 \pm 0.0006$ & Skip loss +0.001 \\
          Dice + CE + PAlign + Skip & Sigmoid & 30 & $0.8873 \pm 0.0005$ & Multi-loss stable \\
          Dice + CE + Skip & LeakyReLU & 30 & $0.8871 \pm 0.0003$ & LeakyReLU stable \\
          Dice + CE + PAlign + Skip & LeakyReLU & 30 & $0.8867 \pm 0.0006$ & Multi-loss stable \\
          Dice + CE + PAlign & ReLU & 11 & $0.8855 \pm 0.0041$ & Phase loss null ($-0.0003$) \\
          \bottomrule
        \end{tabular}
      \end{sc}
    \end{small}
  \end{center}
  \vskip -0.1in 
\end{table}

\begin{table}[ht] 
  \caption{Data Efficiency (Robustness to Training Set Size). \textit{DnCNN} degrades only 1.71\% at 10\% data; \textit{Polar phase-aware with learned gating} degrades 5.29\% ($3.1\times$ worse). Phase-aware is data-hungry due to $\sim$100K learned parameters requiring supervision. Suitable for $N > 100$; unsuitable for $N < 50$.} 
  \label{tab:prel-seg}
  \begin{center}
    \setlength{\tabcolsep}{2.5pt}
    \begin{small}
      \begin{sc}
        \begin{tabular}{lcccccc}
          \toprule
          Data \% & DnCNN PSNR & (ours) PSNR & Gap & DnCNN Degrad & (ours) Degrad & Robustness Ratio \\
          \midrule
          10\% & $28.77 \pm 0.24$ & $25.96 \pm 0.07$ & $-2.81$ & 1.71\% & 5.29\% & 3.1$\times$ \\
          25\% & $28.98 \pm 0.11$ & $26.96 \pm 0.05$ & $-2.02$ & 0.99\% & 1.64\% & 1.7$\times$ \\
          50\% & $28.88 \pm 0.02$ & $26.75 \pm 0.19$ & $-2.13$ & 1.34\% & 2.41\% & 1.8$\times$ \\
          100\% & $29.27 \pm 0.01$ & $27.41 \pm 0.13$ & $-1.86$ & — & — & — \\
          \bottomrule
        \end{tabular}
      \end{sc}
    \end{small}
  \end{center}
  \vskip -0.1in 
\end{table}

\section{Extended Qualitative Visualizations}
\label{app:ext-qual-viz}

To verify that phase-edge alignment is a general property of our architecture and not an artifact of specific image topologies, we provide extended visualizations across distinct semantic categories: Landscapes (high texture), Animals (complex curves), and Sharp Edges (high-contrast geometric boundaries).

\begin{figure}[ht]
  \vskip 0.2in
  \begin{center}
    \centerline{\includegraphics[width=\textwidth]{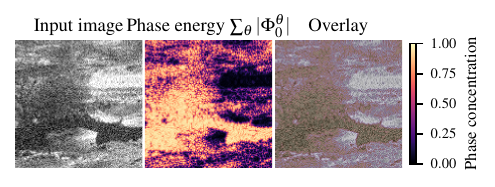}}
    \caption{
      \textbf{Phase energy overlay $\sum_\theta |\Phi^\theta_0|$ for \textit{landscape} image.} Phase concentration is highest at image edges and boundaries, varying across images. This confirms that phase encodes image-specific geometric structure rather than wavelet-intrinsic patterns (see also Figures \ref{fig:phase-enc-land}–\ref{fig:phase-enc-animal})
    }
    \label{fig:phase-overlay-land}
  \end{center}
\end{figure}

\begin{figure}[ht]
  \vskip 0.2in
  \begin{center}
    \centerline{\includegraphics[width=\textwidth]{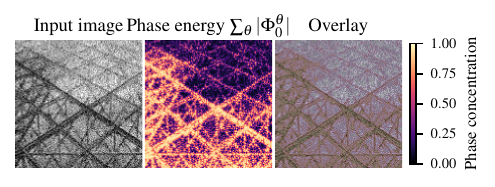}}
    \caption{
      \textbf{Phase energy overlay $\sum_\theta |\Phi^\theta_0|$ for \textit{sharp edges} image.} Phase concentration is highest at image edges and boundaries, varying across images. This confirms that phase encodes image-specific geometric structure rather than wavelet-intrinsic patterns (see also Figures \ref{fig:phase-enc-land}–\ref{fig:phase-enc-animal})
    }
    \label{fig:phase-overlay-sharp}
  \end{center}
\end{figure}

\begin{figure}[ht]
  \vskip 0.2in
  \begin{center}
    \centerline{\includegraphics[width=\textwidth]{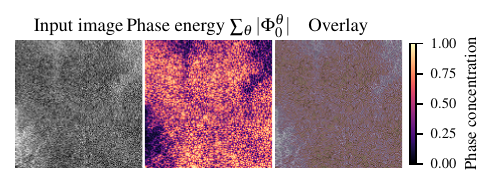}}
    \caption{
      \textbf{Phase energy overlay $\sum_\theta |\Phi^\theta_0|$ for \textit{animal portrait} image.} Phase concentration is highest at image edges and boundaries, varying across images. This confirms that phase encodes image-specific geometric structure rather than wavelet-intrinsic patterns (see also Figures \ref{fig:phase-enc-land}–\ref{fig:phase-enc-animal})
    }
    \label{fig:phase-overlay-animal}
  \end{center}
\end{figure}

\begin{figure}[ht]
  \vskip 0.2in
  \begin{center}
    \centerline{\includegraphics[width=\textwidth]{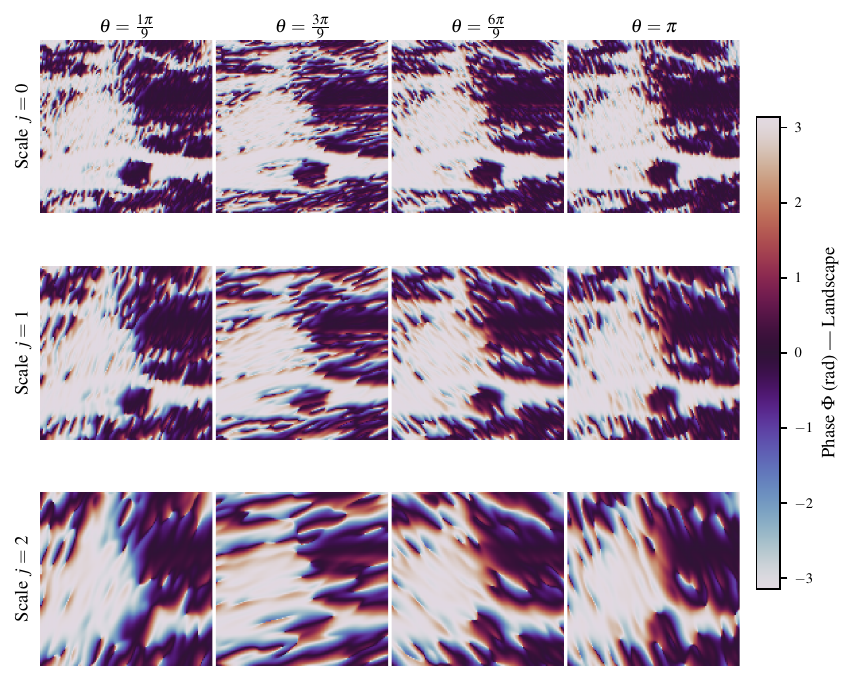}}
    \caption{
        \textbf{Phase maps $\Phi^\theta_j$ at scales $j \in \{0,1,2\}$ and orientations $\theta \in \{1\pi/9, 3\pi/9, 6\pi/9, \pi\}$ for \textit{Landscape}.} Each orientation selectively highlights edges aligned with that direction. Phase structure varies significantly across scales and across images, confirming image-dependent encoding.
    }
    \label{fig:phase-enc-land}
  \end{center}
\end{figure}

\begin{figure}[ht]
  \vskip 0.2in
  \begin{center}
    \centerline{\includegraphics[width=\textwidth]{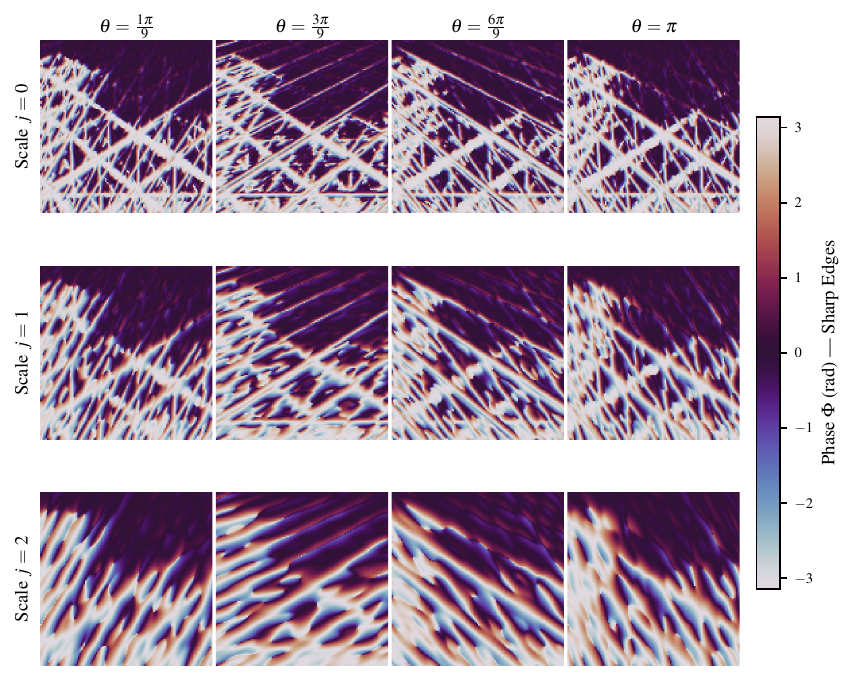}}
    \caption{
        \textbf{Phase maps $\Phi^\theta_j$ at scales $j \in \{0,1,2\}$ and orientations $\theta \in \{1\pi/9, 3\pi/9, 6\pi/9, \pi\}$ for \textit{Sharp Edges}.} Each orientation selectively highlights edges aligned with that direction. Phase structure varies significantly across scales and across images, confirming image-dependent encoding.
    }
    \label{fig:phase-enc-sharp}
  \end{center}
\end{figure}

\begin{figure}[ht]
  \vskip 0.2in
  \begin{center}
    \centerline{\includegraphics[width=\textwidth]{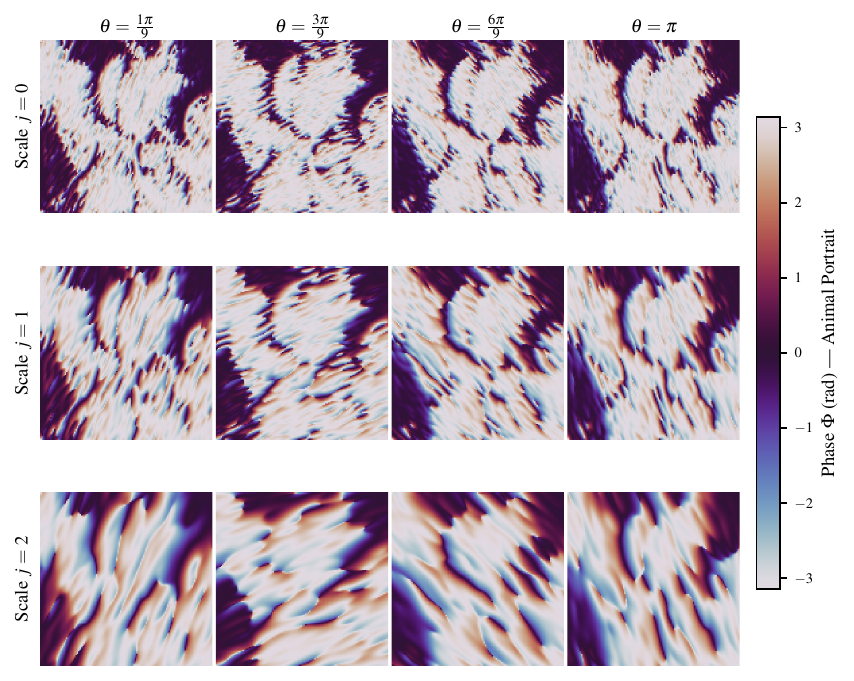}}
    \caption{
        \textbf{Phase maps $\Phi^\theta_j$ at scales $j \in \{0,1,2\}$ and orientations $\theta \in \{1\pi/9, 3\pi/9, 6\pi/9, \pi\}$ for \textit{Animal Portrait}.} Each orientation selectively highlights edges aligned with that direction. Phase structure varies significantly across scales and across images, confirming image-dependent encoding.
    }
    \label{fig:phase-enc-animal}
  \end{center}
\end{figure}

\begin{figure}[ht]
  \vskip 0.2in
  \begin{center}
    \centerline{\includegraphics[width=\textwidth]{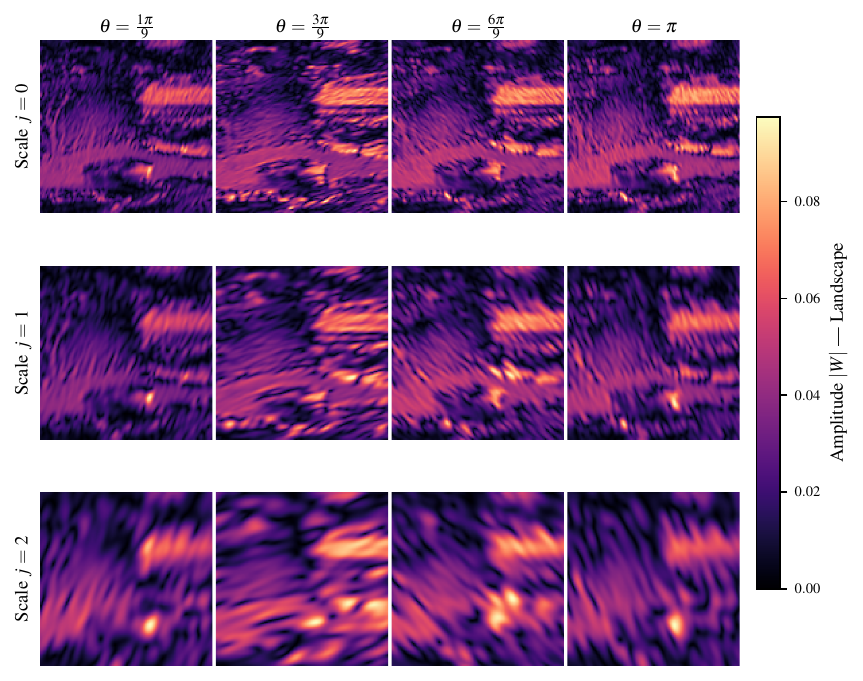}}
    \caption{
        \textbf{Amplitude maps $|W^\theta_j|$ at scales $j \in \{0,1,2\}$ and orientations $\theta \in \{1\pi/9, 3\pi/9, 6\pi/9, \pi\}$ for \textit{Landscape}.} Amplitude encodes directional energy and is smoother and less spatially localized than the corresponding phase maps (Figures \ref{fig:phase-enc-land}–\ref{fig:phase-enc-animal}), consistent with the 5.14× importance ratio in Table~\ref{tab:exp-shuffle}.
    }
    \label{fig:mag-enc-land}
  \end{center}
\end{figure}

\begin{figure}[ht]
  \vskip 0.2in
  \begin{center}
    \centerline{\includegraphics[width=\textwidth]{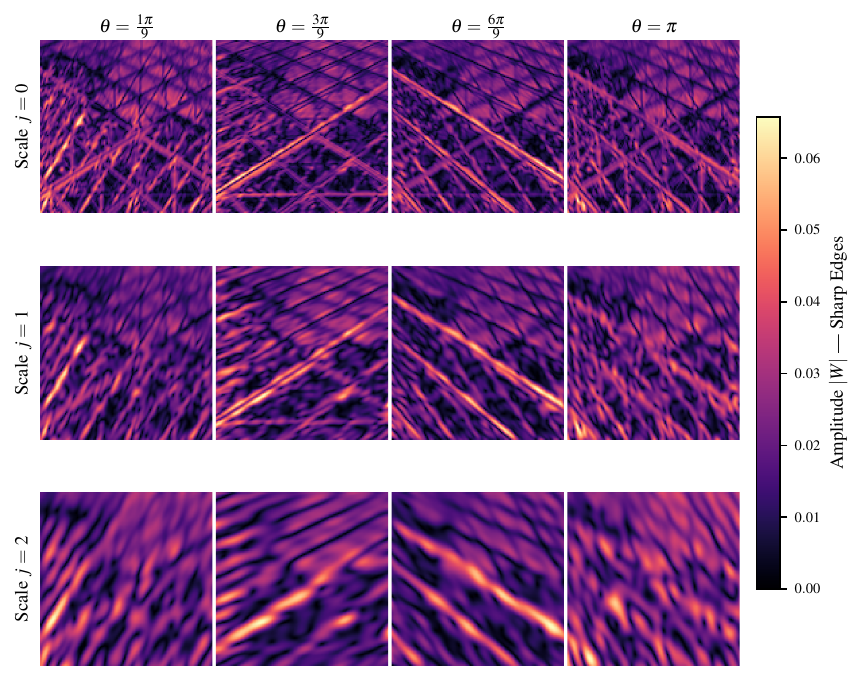}}
    \caption{
        \textbf{Amplitude maps $|W^\theta_j|$ at scales $j \in \{0,1,2\}$ and orientations $\theta \in \{1\pi/9, 3\pi/9, 6\pi/9, \pi\}$ for \textit{Sharp Edges}.} Amplitude encodes directional energy and is smoother and less spatially localized than the corresponding phase maps (Figures \ref{fig:phase-enc-land}–\ref{fig:phase-enc-animal}), consistent with the 5.14× importance ratio in Table~\ref{tab:exp-shuffle}.
    }
    \label{fig:mag-enc-sharp}
  \end{center}
\end{figure}

\begin{figure}[ht]
  \vskip 0.2in
  \begin{center}
    \centerline{\includegraphics[width=\textwidth]{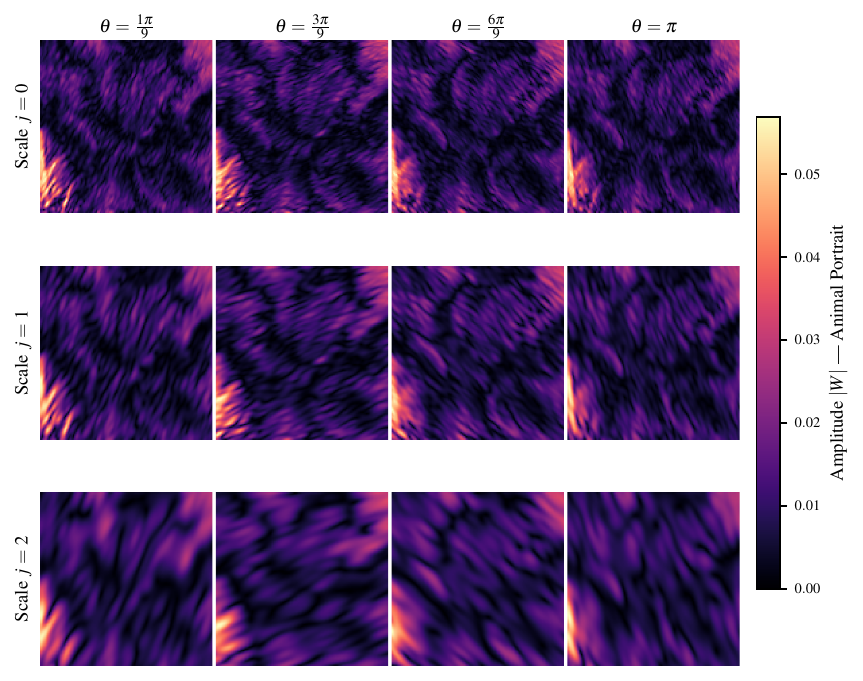}}
    \caption{
        \textbf{Amplitude maps $|W^\theta_j|$ at scales $j \in \{0,1,2\}$ and orientations $\theta \in \{1\pi/9, 3\pi/9, 6\pi/9, \pi\}$ for \textit{Animal Portrait}.} Amplitude encodes directional energy and is smoother and less spatially localized than the corresponding phase maps (Figures \ref{fig:phase-enc-land}–\ref{fig:phase-enc-animal}), consistent with the 5.14× importance ratio in Table~\ref{tab:exp-shuffle}.
    }
    \label{fig:mag-enc-animal}
  \end{center}
\end{figure}

Figures \ref{fig:phase-overlay-land}–\ref{fig:phase-overlay-animal} show phase energy overlays for three image categories. Figures \ref{fig:phase-enc-land}–\ref{fig:phase-enc-animal} show phase maps $\Phi^\theta_j$ across scales and orientations for the same images. Figures \ref{fig:mag-enc-land}–\ref{fig:mag-enc-animal} show the corresponding amplitude maps $|W^\theta_j|$. Comparing phase and amplitude maps across images confirms that phase encodes image-specific geometric structure rather than wavelet-intrinsic patterns.

\end{document}